\documentclass[10pt]{article}
\usepackage[top=0.5in,left=0.8in,footskip=0.75in,marginparwidth=1in]{geometry}
\usepackage{bbm}

\usepackage[utf8]{inputenc}
\usepackage{amsmath}

\usepackage{nameref,hyperref}

\usepackage[right]{lineno}

\usepackage{microtype}
\DisableLigatures[f]{encoding = *, family = * }

\raggedright
\usepackage{changepage}

\usepackage[aboveskip=1pt,labelfont=bf,labelsep=period,singlelinecheck=off]{caption}

\makeatletter
\renewcommand{\@biblabel}[1]{\quad#1.}
\makeatother

\usepackage{lastpage,fancyhdr,graphicx}
\usepackage{epstopdf}
\fancyfootoffset[L]{2.25in}

\usepackage{color}

\definecolor{Gray}{gray}{.25}

\usepackage{graphicx}

\usepackage{sidecap}

\usepackage{wrapfig}
\usepackage[pscoord]{eso-pic}
\usepackage[fulladjust]{marginnote}
\reversemarginpar

\usepackage{float}
\usepackage{ragged2e}
\justifying
 
\usepackage[ruled,vlined]{algorithm2e}

\SetCommentSty{mycommfont}
\SetKwInput{KwInput}{Input}                
\SetKwInput{KwOutput}{Output}              
\newlength\mylen

\usepackage{color,soul}
\usepackage{booktabs, caption, multicol, multirow}
\usepackage{tabularx}
\newcolumntype{?}[1]{!{\vrule width #1}}
\usepackage{amsmath}
\usepackage{amsfonts} 
\usepackage{filecontents}

\begin{document}
\vspace*{0.35in}

\begin{flushleft}
{\Large
\textbf\newline{GRAF: Graph Attention-aware Fusion Networks}
}
\newline
\\
Ziynet Nesibe Kesimoglu\textsuperscript{1},
Serdar Bozdag\textsuperscript{1,2,3,*}
\\
\bigskip
\bf{1} Dept. of Computer Science and Engineering, University of North Texas, Denton, TX
\\
\bf{2} Dept. of Mathematics, University of North Texas, Denton, TX
\\
\bf{3} BioDiscovery Institute, University of North Texas, Denton, TX
\\
\bigskip
* Serdar.Bozdag@unt.edu

\end{flushleft}

\section*{Abstract}

A large number of real-world networks include multiple types of nodes and edges. Graph Neural Network (GNN) emerged as a deep learning framework to generate node and graph embeddings for downstream machine learning tasks. However, popular GNN-based architectures operate on single homogeneous networks. Enabling them to work on multiple networks brings additional challenges due to the heterogeneity of the networks and the multiplicity of the existing associations. In this study, we present a computational approach named GRAF (Graph Attention-aware Fusion Networks) utilizing GNN-based approaches on multiple networks with the help of attention mechanisms and network fusion. Using attention-based neighborhood aggregation, GRAF learns the importance of each neighbor per node (called \textit{node-level attention}) followed by the importance of association (called \textit{association-level attention}). Then, GRAF processes a network fusion step weighing each edge according to learned node- and association-level attentions. Considering that the fused network could be a highly dense network with many weak edges depending on the given input networks, we included an edge elimination step with respect to edges' weights. Finally, GRAF utilizes Graph Convolutional Network (GCN) on the fused network and incorporates node features on graph-structured data for a node classification or a similar downstream task. To demonstrate GRAF’s generalizability, we applied it to four datasets from different domains and observed that GRAF outperformed or was on par with the baselines, state-of-the-art methods, and its own variations for each node classification task. Source code for our tool is publicly available at \href{https://github.com/bozdaglab/GRAF}{https://github.com/bozdaglab/GRAF}.\\

\begin{multicols}{2}

\section{Introduction}
Graph Neural Network (GNN) emerged as a powerful architecture allowing utilization of node features on graph-structured data \cite{gori2005new, scarselli2008graph, kipf2016semi, velivckovic2017graph}. GNN is a framework to define deep neural networks on graph-structured data generating node representations that depend on the graph structure, as well as node features. Iteratively, every node updates its current embedding by aggregating information from its local neighborhood. Graph Convolutional Network (GCN) is one of the most popular GNNs \cite{kipf2016semi}, which gives equal importance to all neighbor nodes. Inspired from \cite{vaswani2017attention}, attention mechanisms are applied to graph-structured data \cite{velivckovic2017graph}, allowing us to learn the importance of neighbor nodes (called \textit{attention}) in a given network.

GNN-based architectures are mostly applicable to a single homogeneous network (i.e., one network with one type of node and edge). However, real-world networks mostly have multiple associations and they could be heterogeneous (i.e., having more than one node type). One approach to utilize GNN on multiple graph-structured data could be running GNN on each network separately. For instance, MOGONET runs multiple GCNs and integrates label distributions from each to decide the final label of nodes \cite{wang2021mogonet}. SUPREME generates node embeddings from each network using GCN and trains a machine learning (ML) model using these embeddings for a downstream task \cite{kesimoglu2022supreme}. Both MOGONET and SUPREME do not utilize graph attentions, however, SUPREME integrates embeddings from each combination of networks to find the best model among all. Nevertheless, running each combination with single networks could be computationally expensive, especially when we have a high number of networks to integrate.

Another approach could be fusing the networks into one and  running GNN on the fused network. However, how to fuse the networks is challenging, especially without knowing the quality of the given networks such as which networks are more important and which node neighbors have stronger relations. Considering advances on graph-structured data, further utilization of GNN on multiple networks could be achieved by improving network fusion strategy with attention mechanisms. 

In this study, we introduce a computational approach named GRAF (Graph Attention-aware Fusion Networks) as a GNN-based approach utilizing attention mechanisms and network fusion simultaneously. To this end, first, we generated meta-path-based networks if the given network is heterogeneous or obtained similarity networks if homogeneous. After that, we obtain node-level and association-level attention following \cite{wang2019heterogeneous}. Node-level attention helps us learn the importance of each node to its neighbors, while association-level attention gives an overall interpretation of the impact of each association type. Combining both node-level and association-level knowledge, we weigh each association and fuse the networks having rich information about the associations. On the fused network, we utilize GCN for a node classification
or a similar downstream task incorporating all the node features that we might have plenty of.

The contributions of our work are summarized as follows:
\begin{itemize}
\item We developed a novel GNN-based approach called GRAF with an attention-aware network fusion strategy. GRAF operates on multiple networks where the associations are properly weighted utilizing attention mechanisms. GRAF runs GCN on the fused network for the desired node classification or a similar downstream task.
\item  GRAF's network fusion step enables us to utilize GNN-based model on a single network, which is enriched with multiple associations according to their importance. Our edge weighting considers the importance of each association and each neighbor node, thus prioritizing edges in a network properly. This increases the interpretability of the networks.
\item We conducted extensive evaluations to measure the performance of our model. Our results suggest that the utilization of attention mechanism and network fusion strategy simultaneously could have superior performance as compared to the state-of-the-art and baseline methods. We applied GRAF to four node classification problems from different domains, showing generalizibility of GRAF.
\end{itemize}

\section{Related Work}
\subsection{GNN methods}
GNN attracted high interest as a deep learning framework to generate node and graph embeddings. Several GNN-based architectures have been developed with different approaches to feature aggregation from the local structure \cite{xu2018powerful, wu2019simplifying, kipf2016semi, velivckovic2017graph}. GCN \cite{kipf2016semi} uses self edges in the neighborhood aggregation and normalizes across neighbors with equal importance \cite{kipf2016semi}. On the cancer type prediction problem, in \cite{ramirez2020classification}, the authors leverage GCN on a single biological network with one data modality, thus limiting the utilization of multiple data and networks. In \cite{rhee2017hybrid}, the authors leveraged GCN and relation network with PPI network on the classification task, while in \cite{zitnik2018modeling}, the authors used a GCN-based model on drug and protein interaction network and separately dealt with different edge types. 

Generalizing the self-attention mechanisms of transformers \cite{vaswani2017attention}, Graph Attention Networks (GAT) has been developed using attention-based neighborhood aggregation learning the importance of each neighbor \cite{velivckovic2017graph}. A follow-up study has shown that GAT computes static attention having the same ranking for attention coefficients in the same graph, and has proposed GATv2 \cite{brody2021attentive} by changing the order of operations, and improved the expressiveness of GAT. 

\subsection{GNN methods on multiplex networks}

To utilize more knowledge, studies improved GNN-based architectures to operate on multiple networks \cite{wang2021mogonet, kesimoglu2022supreme}. For instance, MOGONET runs three different GCNs on three patient similarity networks from three different data modalities separately \cite{wang2021mogonet}. Then, it uses the label distribution from three different models and utilizes them to predict the final label of each node. SUPREME \cite{kesimoglu2022supreme} is introduced as a GCN-based node classification framework operating on multiple networks, encoding multiple features on multiple networks. As compared to MOGONET, SUPREME utilized the intermediate embeddings and integrated them with node features, resulting in a consistent and improved performance. Even though both methods do not utilize graph attention, SUPREME integrates embeddings trying all combinations of networks to find the best model.

Heterogeneous Graph Attention Network (HAN) applies GNN-based architecture on a heterogeneous network utilizing attention mechanisms \cite{wang2019heterogeneous}. A meta-path is defined as a series of relations between nodes defining a new connection between its starting and ending node types. HAN generates meta-path-based networks from the heterogeneous network. Then it applies one transformation matrix for each node type, and learns node-level attention (for each node using its meta-path-based neighborhood) and association-level attention (for each meta-path for the prediction task). Then HAN optimizes the model via backpropagation. Even though this approach gives us a good approximation of the importance of given associations, the hierarchical attention strategy could limit the utilization of data and networks.

\subsection{Network fusion methods}
Since multiple networks may contain complementary information, some studies integrated networks into one network \cite{wang2014similarity, ma2017integrate}. For instance, Similarity Network Fusion (SNF) \cite{wang2014similarity} builds a patient similarity network based on each data modality, fuses all networks into one consensus network by applying a nonlinear step, and performs the clustering on that consensus network. Affinity Network Fusion (ANF) \cite{ma2017integrate} builds upon SNF by simplifying the computational operations needed. Network fusion methods show good performance without using probabilistic modeling, however, these tools highly depend on building a similarity network to integrate information from multiple data modalities. In addition, these tools can not utilize node features on the network, which could be potentially informative.

\section{GRAF}\label{sec:methods}
\begin{figure*}
  \centering
  \includegraphics[width=\linewidth]{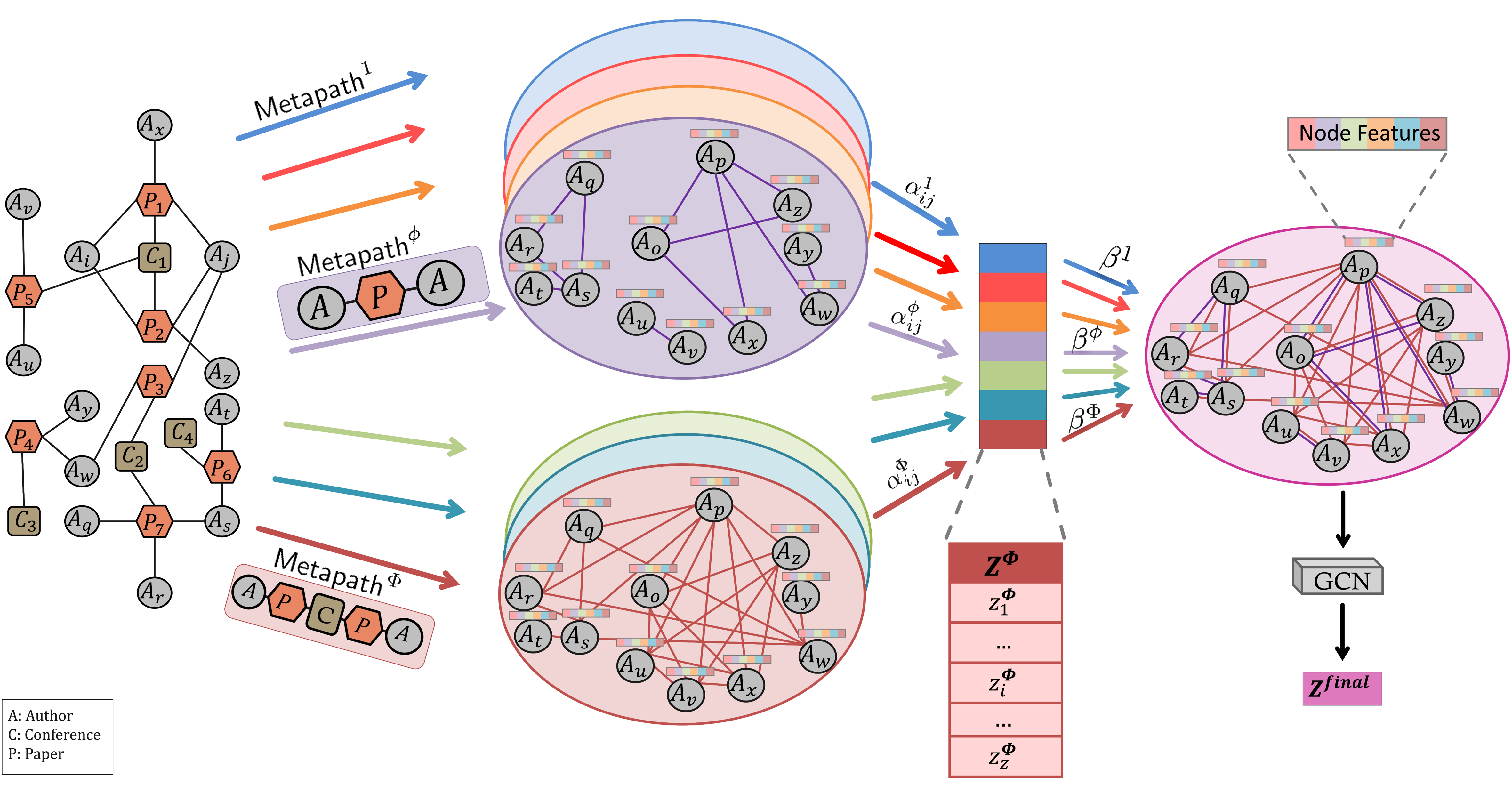}
  \caption{GRAF pipeline on a heterogeneous network. GRAF first generates meta-path-based neighborhood. Then, it obtains node-level and association-level attention. Utilizing those attentions, GRAF fuses multiple networks into one weighted network. GRAF excludes low-weighted edges and learns the node embeddings using graph convolutions on the fused network.}
 \label{gfig:01}
\end{figure*}

GRAF is a computational approach operating graph convolutions on multiple graph-structured data with the utilization of attention mechanisms and network fusion simultaneously (Figure \ref{gfig:01}). Briefly, the first step is data collection and neighborhood generation. In the second step, we obtain node-level and association-level attention. In the third step, GRAF fuses multiple networks into one weighted network utilizing node-level and association-level attention mechanisms. In the last step, GRAF learns the node embeddings using GCN with the convolutions on the fused network. In the following section, we explain each step of GRAF in detail.

\subsection{Neighborhood generation}

When the input network is heterogeneous (IMDB, ACM, and DBLP data for our case), we generated a series of relations between nodes to define a neighborhood between the starting and ending nodes. This type of network generated as a result of a meta-path is called a meta-path-based network and they are highly utilized on heterogeneous networks. We generated meta-path-based networks between the same node type, which is the node of interest in the downstream task. Let's assume that we start with a set of $n$ nodes denoted by $\mathcal{V}$, i.e., $\mathcal{V} = \{v_1, v_2, ..., v_n\}$.

We represented a set of edges (including self edges) for each meta-path-based network with $\mathcal{E}^{\phi}$. The edge $(v_i,v_j) \in \mathcal{E}^{\phi}$ when $v_i$ and $v_j$ have an association based on the meta-path $\phi$ where $\phi \in \{1,2...\Phi\}$ and $\Phi$ is the total number of associations. This could be represented with an indicator function $I$ as follows:
\[
I_{\mathcal{E}^{\phi}}(v_i, v_j)=\left\{\begin{array}{lr}
1 & \text { if }\left(v_{i}, v_{j}\right) \in \mathcal{E}^{\phi}
\\
0 & \text { otherwise }
\end{array}\right.
\]

For a given homogeneous network (DrugADR data for our case), we generated similarity networks based on multiple associations. Similar to the above definition, we represent an edge $(v_i,v_j) \in \mathcal{E}^{\phi}$ when $v_i$ and $v_j$ are associated based on the association $\phi$. 

After generating all $\mathcal{E}^{\phi}$'s for a dataset, we will have a graph $ \mathcal{G}= (\mathcal{V},\mathcal{E}^{\phi})$. In our case, for all datasets, graph $\mathcal{G}$ is an undirected graph, that is, $(v_i,v_j) \in \mathcal{E}^{\phi} \iff (v_j,v_i) \in \mathcal{E}^{\phi}$. We represented the neighborhood as $\mathcal{N}^{\phi}_{i}=\left\{v_{j}:\left(v_{i}, v_{j}\right) \in \mathcal{E}^{\phi}\right\}$, which is the set of nodes having an association with node $v_i$ based on the association $\phi$. We also generated a feature matrix $\mathcal{X} \in \mathbb{R}^{nxf}$ where ${x_i} \in \mathbb{R}^f$ represents the original node features of $v_i$ and $f$ is the input feature size. $\mathcal{X}$ will be used as input for the attention model and final GCN model. 

\subsection{Computing node- and association-level attention}

To learn the importance of each neighbor per association and the impact of each association on the given prediction task, we extracted all the node- and association-level attention values from end-to-end HAN architecture \cite{wang2019heterogeneous}. The following details of this section follow directly from that study. However, these attention values could be obtained from any other approaches.

\paragraph{Node-level attention.} We generated neighborhood connecting nodes through some defined meta-paths or some determined similarities, and supposedly, not all the neighbors connected by that association will have the same effect. Considering that, each node requires weighing its neighbors and giving more importance to highly-attended ones. Inspired from \cite{velivckovic2017graph}, to get the importance of each neighbor, we learned node-level attention. Node-level attention preserves the asymmetry so that the importance of node $v_i$ to $v_j$ is not necessarily the same as the importance of node $v_j$ to $v_i$. In this section, we used the word ``directed'' to show that the value is specific to an edge in the given order. 

First, we did the linear transformation of node features as ${h_i} = {{M_{\Theta}}}.{ x_i}$ where $x_i$ is the original feature for node $v_i$ and $h_i$ is the corresponding projected features. For simplicity, we have one node type $\Theta$ where its transformation matrix is represented with ${M_{\Theta}}$, however, for the heterogeneous networks with more than one node type, we will have different transformation matrices. 

To aggregate the embeddings of neighbors as a weighted average, we learned the attention of each node. To this end, we have $e$, which computes a score for each edge. Since we have multiple associations, and we will have different node-level attention for each association, we represented multiple scoring for each corresponding association. For an edge $(v_i,v_j)$, $e_{ij}^{\phi}$ represents the importance of features of neighbor $v_j$ to the node $v_i$ based on the association $\phi$, i.e., $e_{ij}^{\phi} = \text{LeakyReLU}\left(({{a}}^{\phi})^{\mathbb{T}}.{[h_i||h_j]}\right)$ where ${\bf{a}}^{\phi}$ is node-level attention vector for the association $\phi$, $\mathbb{T}$ is transpose, and $||$ is vector concatenation. Then we normalized the attention scores across neighbors using softmax and obtained the node-level attention $\alpha_{ij}^{\phi}$. The attention of neighbor $v_j$ to the node $v_i$ based on the association $\phi$ for the edge $(v_i,v_j)$ is computed as:
\[\alpha_{ij}^{\phi}=\text{softmax}_j(e_{ij}^{\phi}) = \dfrac{\text{exp}(e_{ij}^{\phi})}{\sum_{k\in \mathcal{N}^{\phi}_{i}}\text{exp}\left(e_{ik}^{\phi}\right)}\]

Utilizing attention weights, we calculated embedding of node $v_i$, $z_i^{\phi}$ as weighted average of embeddings of its neighbors followed by a nonlinear activation function, $\sigma$.
\[z_i^{\phi}=\sigma\left(\sum_{v_j \in \mathcal{N}^{\phi}_{i}}\alpha_{ij}^{\phi}.{ h_j}\right)\]

Considering high variance in the data, multi-head attention is highly preferred for stabilizing the training. Multi-head attention is to repeat the node embedding generation multiple times and to concatenate the embeddings. Following \cite{velivckovic2017graph}, we preferred eight attention heads, thus, we concatenated eight embeddings to generate association-specific embeddings $z_i^{\phi}$. For simplicity, we did not involve multi-head attention in our formulation. Combining all the association-specific node embeddings, we get an association-specific embedding matrix $\mathcal{Z}^{\phi} \in \mathbb{R}^{nxf'}$ where $f'$ is the embedding size.

\noindent\textbf{Association-level attention.} We utilize multiple associations to capture complementary information, however, all the associations might not have a similar impact on the desired task. Therefore, we learn the importance of each association type as follows.

Following \cite{wang2019heterogeneous}, association-specific embeddings are first transformed through a nonlinear transformation and the importance of each association is computed as the similarity of transformed embedding with an association-level attention vector, ${\bf q}$. Then applying average over all nodes in the association $\phi$, $f^{\phi}$ is computed as:
\[f^{\phi} = \dfrac{1}{|\mathcal{V}|}. \sum_{v_i\in \mathcal{V}} {{q}}^{\mathbb{T}}.\text{tanh}({{M_0}}.{ z_i^{\phi}})
\]
These parameters are shared for all the association and association-specific node embeddings. Then normalization happens across all $f^{\phi}$'s using softmax. The association-level attention $\beta^{\phi}$ is:
\[
\beta^{\phi}=softmax_\phi(f^{\phi})=\dfrac{\text{exp}(f^{\phi})}{\text{exp}\left(\sum_{i\in \Phi}f^{i}\right)}
\]
representing the importance of the association $\phi$ for the prediction task. Since higher $\beta^{\phi}$ means a more important association, final node embedding $\mathcal{Z}$ is generated as weighted average of node embeddings followed by a nonlinear activation function $\sigma$:
\[\mathcal{Z}=\sigma\left(\sum_{i \in \Phi}\beta^{i}.\mathcal{Z}^{i}\right)\]

For the node classification task in this step, cross-entropy was minimized over all labeled nodes between the ground truth and the prediction as our loss function. The model was optimized using backpropagation. Due to the variation in the association-level and node-level attention, we preferred to repeat the whole process multiple times (10 times in our experiments) and used the average attention value in the next step. 

Unlike HAN \cite{wang2019heterogeneous}, which concludes with this prediction, we continued with the learned attention, fused the networks, and performed the node classification afterwards.

\subsection{Attention-aware network fusion}

Each node pair might have multiple associations but with different importance for each association. Some node pairs might have edges from insignificant associations, while some node pairs have only one edge, but from a very strong association. Therefore, we cannot only consider the existence of the edges but also the weights. Combining both node-level and association-level attention, we integrated multiple networks into a fused one, on which we did convolution for the prediction task while leveraging the node features.

Since association-level attention represents importance of that association, and node-level attention shows the importance of directed edge on that association, we computed weights of directed edge from $v_i$ to $v_j$ (that is, $score_{\left(v_{i}, v_{j}\right)}$) as weighted sum of available associations as follows:
\[
score_{\left(v_{i}, v_{j}\right)}=\sum_{\phi \in \{1,2...\Phi\}
}  \left(
\beta^{\phi} \alpha_{ij}^{\phi} I_{\mathcal{E}^{\phi}}(v_{i}, v_{j}) \right)
\]

\noindent These scores were utilized to generate a weighted network, which will be used in the prediction task. Intuitively, we give higher weight to the directed edge if it comes from a relatively more important network. But at the same time, we give higher weights to the directed edge if it has higher attention on networks of similar importance. Thus, the importance of node neighbors and the corresponding associations were taken into account, and this edge scoring allowed us to prioritize all directed edges properly (Supplementary Algorithm 1 for the overall attention-aware network fusion strategy).

\subsection{Edge elimination} 
Our fusion step keeps all the edges coming from multiple networks regardless of their weight. This might cause a highly-dense network with many weak edges, depending on the quality of the input networks. Considering that, we included an edge elimination step, where we eliminated some portion of the edges. 

We used the edge weights as their probability to keep that edge in the network and kept x\% of edges by randomly eliminating based on their probabilities. $x$, the percentage of edges to keep, is a hyperparameter to tune. Intuitively, edges that have low attention and/or that are from less important associations will be eliminated from the fused network, keeping the high-quality ones. Now, the fused network is ready to be utilized in GCN model for the downstream task.

\subsection{Node classification task}
To train the fused network utilizing node features and network topology in a downstream task, we utilized GCN \cite{kipf2016semi} and generated final node embeddings.
The input for a GCN model on a single network with the edge set $\mathcal{E}$ is a feature matrix $\mathcal{X} \in \mathbb{R}^{nxf}$
and the adjacency matrix $\mathcal{A} \in \mathbb{R}^{nxn}$ as:
\[
\mathcal{A}[i,j]=\left\{\begin{array}{lr}
score_{\left(v_{i}, v_{j}\right)} & \text { if }\left(v_{i}, v_{j}\right) \in \mathcal{E}
\\
0 & \text { otherwise }
\end{array}\right.
\]

The iteration process of the model is: 
$\mathcal{H}^{(l+1)}=\sigma\left(\mathcal{D}^{-\frac{1}{2}} \mathcal{A} \mathcal{D}^{-\frac{1}{2}} \mathcal{H}^{(l)} \mathcal{W}^{(l)}\right) $ with $\mathcal{H}^{(0)} = \mathcal{X}$ where
\[\mathcal{D}[i,i]=\sum_{j=1}^{n} \mathcal{A}[i,j],\]

$\mathcal{H}^{(l)}$ and $\mathcal{W}^{(l)}$ are activation matrix and trainable weight matrix of $l^{th}$ layer. Feature aggregation on the local neighborhood of each node is done by multiplying $\mathcal{X}$ by $nxn$-sized scaled adjacency matrix $\mathcal{A}^{'}$ where $\mathcal{A}^{\prime}=\mathcal{D}^{-\frac{1}{2}} \mathcal{A} \mathcal{D}^{-\frac{1}{2}}$.

Using a 2-layer GCN model, we had the forward model giving the output $\mathcal{Z_{\text{final}}} \in \mathbb{R}^{nxc}$ where
\[\mathcal{Z_{\text{final}}}=\operatorname{softmax}\left(\mathcal{A}^{\prime} \operatorname{ReLU}\left(\mathcal{A}^{\prime} \mathcal{X} \mathcal{W}^{(1)}\right) \mathcal{W}^{(2)}\right)\]

with $\mathcal{W}^{(1)} \in \mathbb{R}^{fxf'}$, $\mathcal{W}^{(2)} \in \mathbb{R}^{f'xc}$ and $c$ is unique number of class labels. The loss function was calculated by cross-entropy error as in the attention model. 

See Supplementary Methods 1.1 and 1.2 for experimental details.

\section{Experiments}

\subsection{Datasets}
 We applied our tool to four prediction tasks: movie genre prediction from IMDB data (\href{https://www.imdb.com}{https://www.imdb.com}), paper type prediction task from ACM data (\href{http://dl.acm.org}{http://dl.acm.org}), author research area prediction task from DBLP data (\href{https://dblp.uni-trier.de/}{https://dblp.uni-trier.de/}), and drug ADR (adverse drug reaction) prediction task. A detailed description of each dataset is shown in Table \ref{tab:features}.

\begin{table*}[ht]
  \caption{Datasets. [*A: Author, C: Conference, D: Director, M: Movie, P: Paper, R: Actor, S: Subject, T: Term. G-G$_{x}$ denotes drug-drug similarity networks based on four similarities in the order mentioned in Supplementary Methods 1.3.]
}
  \label{tab:features}
  \centering
  \begin{tabular}{lrrrrr}
    \hline
    \textbf{Dataset} & \textbf{\# Samples}& \textbf{\# Features}& \textbf{\# Classes} & \textbf{Association}*&\textbf{\# Edges} \\
    \hline
    \multirow{2}*{IMDB} & \multirow{2}*{4,278} & \multirow{2}*{3,066} & \multirow{2}*{3} & MRM & 85,358\\
    &  &  & & MDM & 17,446\\
    \hline
    \multirow{2}*{ACM} & \multirow{2}*{3,025} & \multirow{2}*{1,870} & \multirow{2}*{3} & PAP & 29,281\\
    &  &  & & PSP & 2,210,761\\

    \hline
    \multirow{4}*{DBLP} & \multirow{4}*{4,057} & \multirow{4}*{334} & \multirow{4}*{4} & APA &  11,113\\
    &  &  & & APAPA & 40,703\\
    &  &  & & APCPA& 5,000,495\\
    &  &  & & APTPA &7,043,627 \\
        \hline
    \multirow{4}*{DrugADR} & \multirow{4}*{664} & \multirow{4}*{1024
    } & \multirow{4}*{5} & G-G$_{1}$ &  8,158\\
    &  &  & & G-G$_{2}$ & 10,518\\
    &  &  & & G-G$_{3}$ & 7,328\\
    &  &  & & G-G$_{4}$ & 3,512 \\
    \hline
  \end{tabular}  
\end{table*}

\noindent\textbf{IMDB:} For movie genre prediction task, we collected and processed IMDB data from PyTorch Geometric library \cite{fey2019fast}. In this data, there are three node types: movie (M), actor (R), and director (D) along with two associations: movie-actor and movie-director. We generated two meta-path-based networks: MRM and MDM. The movies have three genre classes: action, comedy, and drama. Movie node features (which are the elements of a bag-of-words) are from the library's data processing.

\noindent\textbf{ACM:} For paper type prediction task, we collected ACM data using Deep Graph Library \cite{wang2019dgl}. There are three node types: paper (P), author (A), and subject (S). Two associations exist in the data: paper-author and paper-subject. We extracted two meta-path-based networks: PAP and PSP. Paper class labels correspond to three areas: database, wireless communication, and data mining. Paper node features are the elements of a bag-of-words representation, obtained from the library.

\noindent\textbf{DBLP:} For author research area prediction task, we collected DBLP data from \cite{ji2010graph} and preprocessed data using \cite{fu2020magnn}. In this data, there are four node types: paper (P), author (A), conference (C), and term (T) along with three associations: paper-author, paper-conference, and paper-term. We extracted four meta-path-based networks: APA, APAPA, APCPA, and APTPA. Labels of authors correspond to four research areas: database, data mining, artificial intelligence, and information retrieval. Author features are from \cite{wang2019heterogeneous}'s preprocessed data.

\noindent\textbf{DrugADR:} For drug ADR prediction task, we collected drug-ADR pairs from ADReCS \cite{cai2015adrecs}. We got five most frequent ADRs as our class labels: dizziness, hypersensitivity, pyrexia, rash, and vomiting. We collected four drug similarity networks from \cite{olayan2018ddr} and generated SMILES fingerprints as node features (Supplementary Methods 1.3 for details).

\subsection{State-of-the-art and baseline methods}

\textbf{GCN} \cite{kipf2016semi}: Since GCN cannot work on multiple networks, we ran GCN on each association-based network and reported the best performance. Since we utilized GCN in our final prediction, this is also a variant of GRAF, where we do not have any attention mechanism and network fusion.

\noindent\textbf{GAT} \cite{velivckovic2017graph} \textbf{and GATv2} \cite{brody2021attentive}: Since GAT and GATv2 involve attention mechanism aggregating on single homogeneous networks, we could not run it with multiple networks. Instead, we ran them on each association-based network and reported the best performance.

\noindent\textbf{Baseline methods}: We evaluated Multi-layer Perceptron (MLP), Random Forest (RF), and Support Vector Machine (SVM), which use only node features, without utilization of graph-structured data.

\noindent\textbf{HAN} \cite{wang2019heterogeneous}: HAN is an integrative approach operating on multiple networks utilizing attention mechanisms. HAN is also a variant of GRAF without network fusion and edge elimination steps.

\noindent\textbf{SUPREME} \cite{kesimoglu2022supreme}: SUPREME is a GCN-based integrative tool 
generating node embeddings from multiple networks using GCN and training an ML model using them.

\noindent\textbf{$\text{GRAF}_{asc}$}: It is a variant of GRAF including only association-level attention in edge scoring (thus excluding node-level attention). Therefore the score function is replaced with:
\[score_{\left(v_{i}, v_{j}\right)}=
\sum_{\phi \in \{1,2...\Phi\}}  
\left(\beta^{\phi} I_{\mathcal{E}^{\phi}}(v_{i}, v_{j}) \right)
\]
Thus, this variant assigns same importance to every edge in the same association-specific network.

\noindent\textbf{$\text{GRAF}_{node}$}: This variant includes only node-level attention in edge scoring (excluding association-level attention). That is, it assigns the same importance to each association type by replacing the score function with:
\[
score_{\left(v_{i}, v_{j}\right)}=\sum_{\phi \in \{1,2...\Phi\}
}  \left(\alpha_{ij}^{\phi}  I_{\mathcal{E}^{\phi}}(v_{i}, v_{j}) \right)
\]

\noindent\textbf{$\text{GRAF}_{edge}$}:  It is a variant of GRAF, which includes both node-level and association-level attention in edge weighting, however, it does not eliminate edges (i.e., uses all the fused edges).

\subsection{Results}

We evaluated GRAF and the other tools and reported their performance based on three metrics: macro F1 score (macro-averaged F1 score), weighted F1 score (weighted-averaged F1 score), and accuracy. We repeated each run 10 times and included standard deviations with the median of repeats. 

According to our results, GRAF achieved the best performance or was on par with the other tools for all metrics in all datasets (Tables \ref{tab:grafResults} and S1). GRAF, GCN, GAT, GATv2, RF, and SVM had more consistent performance with small standard deviations, while other tools had higher standard deviations, especially in DrugADR data. GRAF outperformed MLP, RF, and SVM showing the importance of graph-structured data utilization. GCN had the best accuracy and weighted F1 in DrugADR data, however GRAF outperformed GCN in terms of macro F1, which is a more suitable metric as ADR labels are not balanced (Supplementary Methods 1.3). 

SUPREME trains a separate ML model for each network combination to find the best network combination. Therefore, SUPREME outputs multiple results instead of one. To make a fair comparison with SUPREME, we reported the worst, median, and best metrics based on validation macro F1 among all combinations, named $\text{SUPREME}_{min}$, $\text{SUPREME}_{med}$, and $\text{SUPREME}_{max}$, respectively. Even though GRAF achieved better than the median of SUPREME results, $\text{SUPREME}_{max}$ was slightly better than GRAF in ACM and DBLP data. On the other hand, for the IMDB and DrugADR data, GRAF outperformed all SUPREME models with a high difference.

\begin{table*}[ht]
  \caption{Node classification results as macro F1 scores (\%)}
  \label{tab:grafResults}
  \centering
  \begin{tabular}{lrrrrr}
    \hline
    \textbf{Method} & \textbf{IMDB}& \textbf{ACM}& \textbf{DBLP} & \textbf{DrugADR}\\
 \hline
GCN&58.7±0&91.5±0&90.5±0&32.9±0\\
GAT&56.8±0&91.0±0&91.4±1&31.6±2\\
GATv2&56.8±1&90.9±1&90.0±1&31.2±2\\
MLP&55.0±1&89.0±1&78.4±1&22.0±4\\
RF&53.4±0&88.9±0&69.3±0&28.8±1\\
SVM&55.1±0&88.5±0&76.5±0&24.8±0\\
HAN&60.9±0&92.0±1&91.5±1&30.2±0\\\midrule
$\text{SUPREME}_{min}$& 53.7±2&90.7±0&77.9±2&31.3±5&\\
$\text{SUPREME}_{med}$& 57.0±2&92.4±1&90.8±1&31.4±4\\
$\text{SUPREME}_{max}$& 60.8±3&\textbf{93.4±1}&\textbf{92.3±2}&32.1±3\\
\hline
$\text{GRAF}_{asc}$&56.3±0&84.6±2&89.7±0&28.8±2\\
$\text{GRAF}_{node}$&\textit{61.3±0}&90.9±2&90.3±1&31.8±2\\
$\text{GRAF}_{edge}$&\textbf{62.1±0}&92.3±0&91.3±1&\textit{33.9±2}\\
\hline
GRAF&\textbf{62.1±0}&\textit{92.6±0}&\textit{91.7±1}&\textbf{34.7±2}\\
   \hline
  \end{tabular}
{
 \begin{flushleft}
 \footnotesize 
Performance evaluation as macro F1 scores in four node classification tasks: movie genre prediction from IMDB data, paper type prediction task from ACM data, author research area prediction task from DBLP data, and drug ADR (adverse drug reaction) prediction task. The best result is highlighted as bold and second-best as italic. $\text{GRAF}_{asc}$ and $\text{GRAF}_{node}$ denote GRAF with only association- and only node-level attention, respectively. $\text{GRAF}_{edge}$ denotes GRAF without edge elimination. $\text{SUPREME}_{min}$, $\text{SUPREME}_{med}$, and $\text{SUPREME}_{max}$ denote the worst, median, and best model based on validation macro F1 among all network combinations. GCN, GAT, and GATv2 were evaluated for every single network, and the best performance was reported. [GAT: Graph Attention Network, GCN: Graph Convolutional Network, MLP: Multi-layer Perceptron, RF: Random Forest, SVM: Support Vector Machine].
 \end{flushleft}}
 \end{table*}

We got input attention values from HAN and utilized them in network fusion step. GRAF outperformed HAN for all prediction tasks, suggesting that our attention-aware network fusion strategy could improve the utilization of multiple graph-structured data further. In all datasets, GRAF had outperformance over GCN, GAT, and GATv2 showing the importance of multiple network utilization. Overall, integrative approaches (i.e., SUPREME, GRAF, and HAN) had better performance. 

We observed that node-level and association-level attentions are crucial for GRAF's performance. Using only association-level attention, $\text{GRAF}_{asc}$ had lower performance for all datasets, which is not surprising since all edges in the same network had the same importance. On the other hand, using only node-level attention, $\text{GRAF}_{node}$ had lower performance than GRAF, but had better performance than $\text{GRAF}_{asc}$. $\text{GRAF}_{node}$ assigned the same importance to each association, but the node-level attention kept a good amount of knowledge. $\text{GRAF}_{edge}$ had comparable performance with GRAF. 

To check GRAF's performance on different-sized data splits, we generated four additional split sets on IMDB data (Supplementary Methods 1.4). In all split sets, GRAF had the best performance among other methods consistently (Figures S1, S2, and S3). We also observed that most methods showed a tendency to increase their performance with higher training sample size, which is expected.

To check the impact of eliminated edges, we compared their performance on all datasets (Figures S4, S5, and S6). For relatively easier tasks like on ACM and DBLP data, we did not see much difference, even between keeping only 10\% of edges vs. no elimination. For tasks like on IMDB and DrugADR data, we observed that the kept edge percentage does make a difference
. In IMDB data, hyperparameter tuning resulted in no edge elimination, thus GRAF and $\text{GRAF}_{edge}$ results were the same. However, the other datasets utilized edge elimination (specifically keeping 70\%, 70\%, and \%30 of the edges for ACM, DBLP, and DrugADR data, respectively).

GRAF allows us to interpret the results of given tasks using node-level attention, association-level attention, and also the fused scores combining both attentions. We reported association-level attention to see which association is more useful in general (Table S2). IMDB data had similar attention, while ACM and DBLP data had one network with strong attention (> 0.6). On these datasets, GCN, GAT, and GATv2 had the highest performance using the networks with the highest attention value. This result was also consistent with HAN's findings \cite{wang2019heterogeneous}. 

Our tool needs attention values as prior to be able to fuse multiple networks. We utilized HAN \cite{wang2019heterogeneous} for this purpose. However, HAN might be computationally expensive, especially when we have big networks and/or many networks to integrate. To address this limitation, the node-level attention could be computed using a more efficient approach such as GAT. To compute association-level attention more efficiently, calculations can be performed on sampled subgraphs instead of the entire network.


\section{Conclusion}
In this study, we developed a computational framework to utilize multiple graph-structured data from homogeneous or heterogeneous networks with attention mechanisms and network fusion. The proposed GRAF showed increased performance over the state-of-the-art and baseline methods and the variants of the tool itself. We applied GRAF to four different datasets from different domains, to show that GRAF is generalizable. Our results showed that GRAF outperformed or had comparable results with other tools on four node classification tasks.



%
%

\vspace*{-10pt}


\section*{Funding}
This work was supported by the National Institute of General Medical Sciences of the National Institutes of Health under Award Number R35GM133657 to SB.

\end{multicols}

\bibliography{bib}

\bibliographystyle{unsrt}
\makeatletter\@input{supp.tex}\makeatother
\end{document}


\vspace*{0.35in}

\begin{flushleft}
{\Large
\textbf\newline{GRAF: GRAPH ATTENTION-AWARE FUSION NETWORKS}
}
\newline
\\
Ziynet Nesibe Kesimoglu\textsuperscript{1},
Serdar Bozdag\textsuperscript{1,2,3}
\\
\bigskip
\bf{1} Dept. of Computer Science and Engineering, University of North Texas, Denton, TX
\\
\bf{2} Dept. of Mathematics, University of North Texas, Denton, TX
\\
\bf{3} BioDiscovery Institute, University of North Texas, Denton, TX
\\
\bigskip

\end{flushleft}

\section{Supplementary Methods}

\subsection{Data splitting}
For ACM, IMDB, and DBLP datasets, we kept the number of sample sizes in training/validation/test splits the same as the used library. ACM data has 600/300/2125 samples for training/validation/test splits, while IMDB data has 400/400/3478 and DBLP data has 400/400/3257, respectively. However, we regenerated training, validation, and test splits as stratified, thus keeping the class label ratio of original data same for each split. For DrugADR data, we divided 664 drugs into training/validation/test splits as stratified and having 338/113/113 (60\%/20\%/20\%) drugs, respectively. 

\subsection{Experimental setup}
We used the training and validation splits to tune the hyperparameters (i.e., hidden layer size, learning rate, and percentage of edges to keep) of the final model. We tried learning rates of 0.01, 0.001, 0.005, and 0.0001; hidden sizes of 32, 64, 128, 256, and 512; and 10 values as hyperparameters of the edge elimination step (namely 10\%, 20\%, ... and 100\%, where x\% means x\% of the edges will be kept based on the normalized weights, and 100\% means no edge elimination). 

We repeated whole process 10 times for each hyperparameter combination and used the hyperparameter combination giving the best median macro-weighted F1 (macro F1) score on the validation data. Using this hyperparameter combination, final model was built and evaluated 10 times on the test data, which was never seen during training and hyperparameter tuning. The evaluation metrics (macro F1 score, weighted F1 score, and accuracy) were obtained from the median of these 10 runs.

Dropout was 0.6, head size was 8, and weight decay was 0.001 for HAN model (as default). For all the models, early stopping was used with patience of 10 and Adam optimization \cite{kingma2014adam} was used to optimize. 

The experiments were run on servers with 2x AMD 7763 2.45GHz CPUs and 2048GB RAM. When we checked computational time on DrugADR data after calculating attention weights (i.e., running HAN), GRAF takes approximately 10 minutes with all edges kept and time reduces when edges were eliminated. For different hidden size and learning rates, computational time varied up to 1 hour. We observed HAN's training time was not consistent taking 10 minutes to 8 hours per run.

\subsection{DrugADR data preparation}

For drug ADR prediction task, we collected drug-ADR pairs from ADReCS (v3.1) \cite{cai2015adrecs}. We got five most frequent ADR (with more than 300 frequencies) using ADR Frequency (FAERS) readily available in ADReCS data. We used them as our class labels and mapped drugs to their most frequent ADRs among these five labels. We ended up with 664 drugs: 192 with Dizziness, 57 with Hypersensitivity, 127 with Pyrexia, 132 with Rash, and 156 with Vomiting.

We collected 1024-bit SMILES fingerprints as node features using RDKit (Open-source cheminformatics at \href{https://www.rdkit.org}{https://www.rdkit.org}) on drug structures obtained from DrugBank \cite{wishart2006drugbank} (released on 2023/01/04 as SDF format). 

We collected four drug similarity networks from \cite{olayan2018ddr} based on: drug ATC (Anatomical Therapeutic Chemical) code-based similarity, drug interactions-based similarity, chemical structures-based molecular fingerprints similarity, and drug side effects-based similarity. All files are under the folder DDR\_data/DrugBank/sim\_line at \href{https://bitbucket.org/RSO24/ddr/src/master/DDR_Data.zip}{https://bitbucket.org/RSO24/ddr/src/master/DDR\_Data.zip}. Files used in this study are listed below:
\begin{itemize}
    \item     All\_DB\_Did\_ATC\_codes\_FDA\_Sim\_D\_T\_in\_R\_basedOn\_Dsmi\_TNoFalseAA\_interacts
    \item All\_DB\_Did\_D\_interactions\_FDA\_Sim\_D\_T\_in\_R\_basedOn\_Dsmi\_TNoFalseAA\_interacts
    \item DRUGBANK\_FDA\_SIMCOMP\_similarity\_filtered\_D\_T\_in\_R\_basedOn\_Dsmi\_TNoFalseAA\_interacts
    \item DrugBank\_SIDER\_Sim\_D\_T\_in\_R\_basedOn\_Dsmi\_TNoFalseAA\_interacts
\end{itemize}
The details of each similarity network are in the Supplemental file of \cite{olayan2018ddr}. We collected all the networks and checked their data distributions. These networks were weighted and in range (0,1], and a high number of interactions exist in each. We only kept the edges where the weight is $1$, which is the maximum possible value. If we end up with an edge number less than 1000, then we decreased the threshold by $0.1$. Finally, we had 8,158 edges in drug ATC code-based similarity network (with threshold $\ge 0.6$), 10,518 edges in drug interactions-based similarity network (with threshold $\ge 0.6$), 7,328 edges in chemical structures-based molecular fingerprints similarity network (with threshold $\ge 0.4$), and 3,512 edges in drug side effects-based similarity network (with threshold $\ge 0.3$). 


\subsection{Sensitivity to different data splits}
To check the impact of different training percentages on prediction results, we generated additional split sets with different training/validation percentages. We kept 20\% of data for testing only and this split was the same for all the split sets. We generated four additional split sets with the thresholds 20\%, 40\%, 60\%, and 80\%. Threshold x\% means that x\% of the remained data (i.e. data excluding test split) is used as training split while using the rest as validation. Thus, training/validation/test splits have 20\%/16\%/64\%, 20\%/32\%/48\%, 20\%/48\%/32\%, and 20\%/64\%/16\% for the thresholds 20\%, 40\%, 60\%, and 80\%, respectively. We did this analysis on IMDB data only, since it is the smallest dataset with only two networks. For other data with four networks and with big networks (millions of edges), we had increased computational time, especially when we have higher sample size in training split. Similarly, we excluded SUPREME from this analysis because of increased computational time. We repeated each evaluation 10 times for each split set and reported the median of those runs in Supplemental Figures S1, S2, and S3.

\newpage

\newpage
\section{Supplementary Algorithms}
The overall attention-aware network fusion strategy is shown in Supplementary Algorithm \ref{alg:netwFusion}. Bias vectors before applying the non-linearity are omitted for simplicity.

\begin{algorithm}[h]
\DontPrintSemicolon
  \setlength{\lineskip}{5pt}
  \KwInput{Graph $ \mathcal{G}= (\mathcal{V},\mathcal{E})$ where $\mathcal{V}$ is a set of $n$ nodes, i.e., $\mathcal{V} = \{v_1, v_2, ..., v_n\}$, and $\mathcal{E}^{\phi}$ is a set of edges between nodes \newline
  $(v_i,v_j) \in \mathcal{E}^{\phi}$ when $v_i$ and $v_j$ have based on the association $\phi$ a
  \newline
  $(v_i,v_j) \in \mathcal{E}^{\phi} \iff (v_j,v_i) \in \mathcal{E}^{\phi}$ (undirected graph)\newline
  feature matrix $\mathcal{X} \in \mathbb{R}^{nxf}$ 
  }

  \KwOutput{Adjacency matrix $\mathcal{A}$}
  \For{$c \in \{1,2...C\}$}{
  
    \For{$\phi \in \{1,2...\Phi\}$} 
    {
        Node type-specific transformation: ${h_i} = {{M_{\Theta}}}.{ x_i}$ \;
        \For{$v_i \in \mathcal{V}$}
        {
            \For{$v_j \in \mathcal{N}^{\phi}_{i}=\left\{v_{j}:\left(v_{i}, v_{j}\right) \in \mathcal{E}^{\phi}\right\}$}
            {
                $e_{ij}^{\phi} = \text{LeakyReLU}\left(({{a}}^{\phi})^{\mathbb{T}}.{[h_i||h_j]}\right)$ \;
                Weight coefficient $\alpha_{ij}^{\phi}=$softmax$_j(e_{ij}^{\phi}) = \dfrac{\text{exp}(e_{ij}^{\phi})}{\sum_{k\in \mathcal{N}^{\phi}_{i}}\text{exp}\left(e_{ik}^{\phi}\right)}$ \;
            }
        }
        Association-specific embedding ${z_i^{\phi}}=\sigma\left(\sum_{v_j \in \mathcal{N}^{\phi}_{i}}\alpha_{ij}^{\phi}.{ h_j}\right)$ \;
        
        Association-specific combined embedding ${ \mathcal{Z}^{\phi}}$ \;
        $f^{\phi} = \dfrac{1}{|\mathcal{V}|}. \sum_{v_i\in \mathcal{V}} {{q}}^{\mathbb{T}}.$tanh$({{M_0}}.{ z_i^{\phi}})$  \;
    }
   Association weight coefficient $\beta^{\phi}=$softmax$_\phi(f^{\phi})=\dfrac{\text{exp}(f^{\phi})}{\text{exp}\left(\sum_{i\in \Phi}f^{i}\right)}$ \;

    Final embedding $\mathcal{Z}=\sigma\left(\sum_{i \in \Phi}\beta^{i}.\mathcal{Z}^{i}\right)$ \;
    Cross-entropy loss \;

    Backpropagation and final parameter obtention\;

    $\beta^{\phi}_{c} = \beta^{\phi}$ \; 
    $\alpha_{ijc}^{\phi} = \alpha_{ij}^{\phi}$\;

}
$\beta^{\phi} = \frac{1}{C}\sum_{c \in \{1,2...C\}}\beta^{\phi}_{c}$ \;
$\alpha_{ij}^{\phi} = \frac{1}{C}\sum_{c \in \{1,2...C\}}\alpha_{ijc}^{\phi}$ \;
$
\mathcal{A}[i,j]=\sum_{\phi \in \{1,2...\Phi\}
}  \left(
\beta^{\phi} \alpha_{ij}^{\phi}  I_{\mathcal{E}^{\phi}}(v_{i}, v_{j})
\right)
$
\caption{Attention-aware network fusion.}
\label{alg:netwFusion}\end{algorithm}

\newpage
\section{Supplementary Tables}

\begin{table*}[h]
  \caption{Node classification results as accuracy and weighted F1 scores (\%) [Performance evaluation as accuracy and weighted F1 scores in four node classification tasks: movie genre prediction from IMDB data, paper type prediction task from ACM data, author research area prediction task from DBLP data, and drug ADR (adverse drug reaction) prediction task. The best result is highlighted as bold and second-best as italic. $\text{GRAF}_{asc}$ and $\text{GRAF}_{node}$ denote GRAF with only association- and only node-level attention, respectively. $\text{GRAF}_{edge}$ denotes GRAF without edge elimination. $\text{SUPREME}_{min}$, $\text{SUPREME}_{med}$, and $\text{SUPREME}_{max}$ denote the worst, median, and best model based on validation macro F1 among all network combinations. GCN, GAT, and GATv2 were evaluated for every single network, and the best performance was reported. [GAT: Graph Attention Network, GCN: Graph Convolutional Network, MLP: Multi-layer Perceptron, RF: Random Forest, SVM: Support Vector Machine].]}
  \label{suptab:grafResults}
  \centering
  \begin{tabular}{lrrrr?{0.5mm}rrrr}
    \hline
    & \multicolumn{4} {c} {\textbf {Accuracy}} & \multicolumn{4} {c} {\textbf {Weighted F1}}\\
    \hline
    \textbf{Method} & \textbf{IMDB}& \textbf{ACM}& \multicolumn{1} {c}{\textbf{DBLP}}& \textbf{DrugADR}& \textbf{IMDB}& \textbf{ACM}& \textbf{DBLP}&\textbf{DrugADR}\\
 \hline
GCN&58.5±0&91.5±0&91.2±0&\textbf{41.4±0}&58.5±0&91.5±0&91.4±0&\textbf{43.2±0}\\
GAT&56.9±0&91.0±0&92.1±1&\textit{41.0±2}&56.7±0&91.0±0&92.0±1&37.7±2\\
GATv2&56.8±1&90.9±1&90.8±1&39.8±2&56.6±1&90.9±1&90.7±1&37.6±2\\
MLP&55.0±1&89.0±1&79.2±1&30.4±4&54.8±1&88.9±1&79.2±1&26.2±5\\
RF&53.7±0&89.0±0&70.7±0&39.1±1&53.7±0&88.9±0&70.3±0&34.3±1\\
SVM&54.9±0&88.5±0&77.1±0&26.3±0&55.0±0&88.5±0&77.1±0&26.3±0\\
HAN&61.1±0&91.9±1&92.1±1&38.3±1&60.9±0&91.9±1&92.0±1&35.6±1\\
\hline
$\text{SUPREME}_{min}$&54.5±1&90.7±1&78.6±2&36.5±4&54.2±2&90.7±0&78.6±2& 34.0±5\\
$\text{SUPREME}_{med}$&57.1±1&92.4±1&91.4±1&38.3±3&57.0±2&92.3±1&91.4±1&36.2±3\\
$\text{SUPREME}_{max}$&60.9±2&\textbf{93.4±1}&\textbf{92.9±2}&38.3±2&60.7±3&\textbf{93.3±1}&\textbf{92.9±2}&36.5±2\\
\hline
$\text{GRAF}_{asc}$&56.9±0&84.3±2&90.6±0&34.2±2&57.0±0&84.2±2&90.8±0&35.7±3\\
$\text{GRAF}_{node}$&\textit{61.6±0}&90.8±2&91.0±1&38.3±3&\textit{61.7±0}&90.9±2&91.1±1&40.0±4\\
$\text{GRAF}_{edge}$&\textbf{62.3±0}&92.2±0&91.9±1&40.6±2&\textbf{62.3±0}&92.2±0&91.9±0&41.8±3\\
\hline
GRAF&\textbf{62.3±0}&\textit{92.5±0}&\textit{92.2±1}&\textit{41.0±2}&\textbf{62.3±0}&\textit{92.5±0}&\textit{92.2±1}&\textit{42.0±2}\\

   \hline
  \end{tabular}
\end{table*}

\begin{table*}[ht]
  \caption{Association-level attentions [*A: Author, C: Conference, D: Director, M: Movie, P: Paper, R: Actor, S: Subject, T: Term. G-G$_x$ denotes drug-drug similarity networks based on four similarities in the order mentioned in Supplementary Methods 1.3.]}
  \label{tab:attentions}
  \centering
  \begin{tabular}{lrrrrr}
    \hline
    \textbf{Dataset} &  \textbf{Association}*&\textbf{Attention} \\
    \hline
    \multirow{2}*{IMDB} & MRM & 0.44\\
    & MDM & 0.56\\
    \hline
    \multirow{2}*{ACM}  & PAP & 0.66\\
    & PSP & 0.34\\

     \hline
    \multirow{4}*{DBLP} & APA &  0.23  \\
    & APAPA & 0.13\\
    & APCPA& 0.64\\
    & APTPA & 0.01 \\
        \hline
    \multirow{4}*{DrugADR} & G-G$_{1}$ &  0.19\\
    & G-G$_{2}$ & 0.14\\
    & G-G$_{3}$ & 0.23\\
    & G-G$_{4}$ & 0.43 \\
    
    \hline
  \end{tabular}  
  
\end{table*}

\newpage
\section{Supplementary Figures}

\begin{figure}[h]
  \centering
  \includegraphics[width=0.85\linewidth]{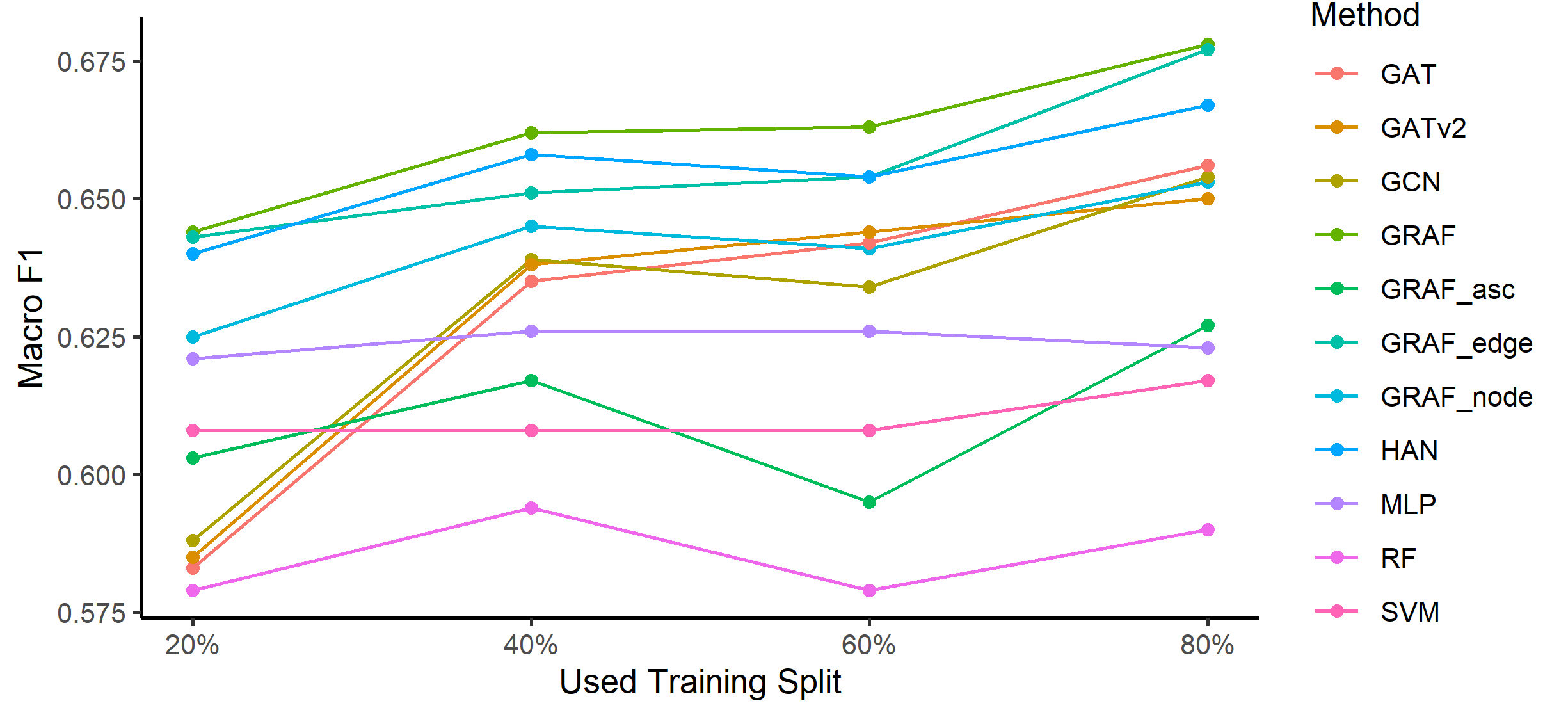}
 \caption{Performance with different training splits on IMDB data (macro F1).}
 \label{gfig:s1_1}
\end{figure}

\begin{figure}[h]
  \centering
  \includegraphics[width=0.85\linewidth]{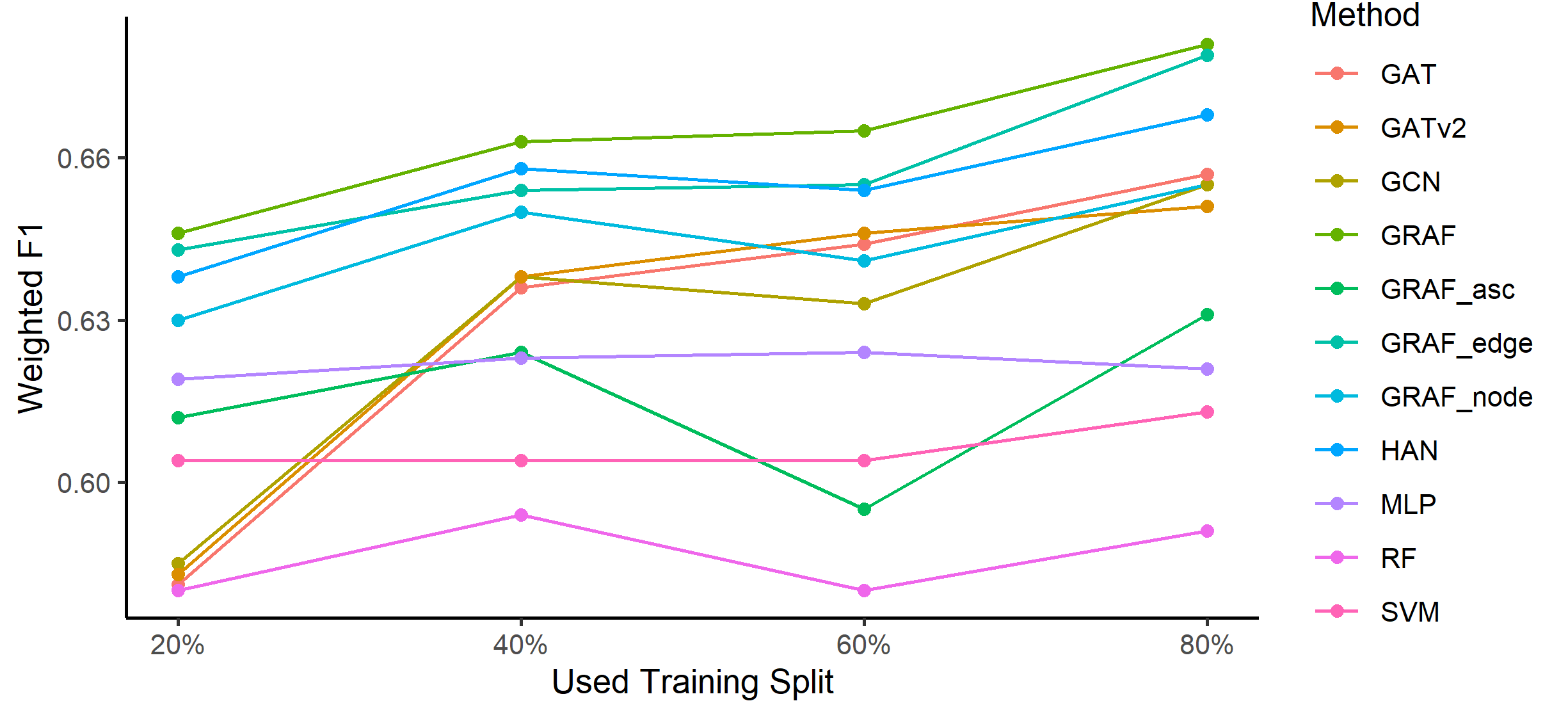}
  \caption{Performance with different training splits on IMDB data (weighted F1).}
 \label{gfig:s1_2}
\end{figure}

\begin{figure}[h]
  \centering
  \includegraphics[width=0.85\linewidth]{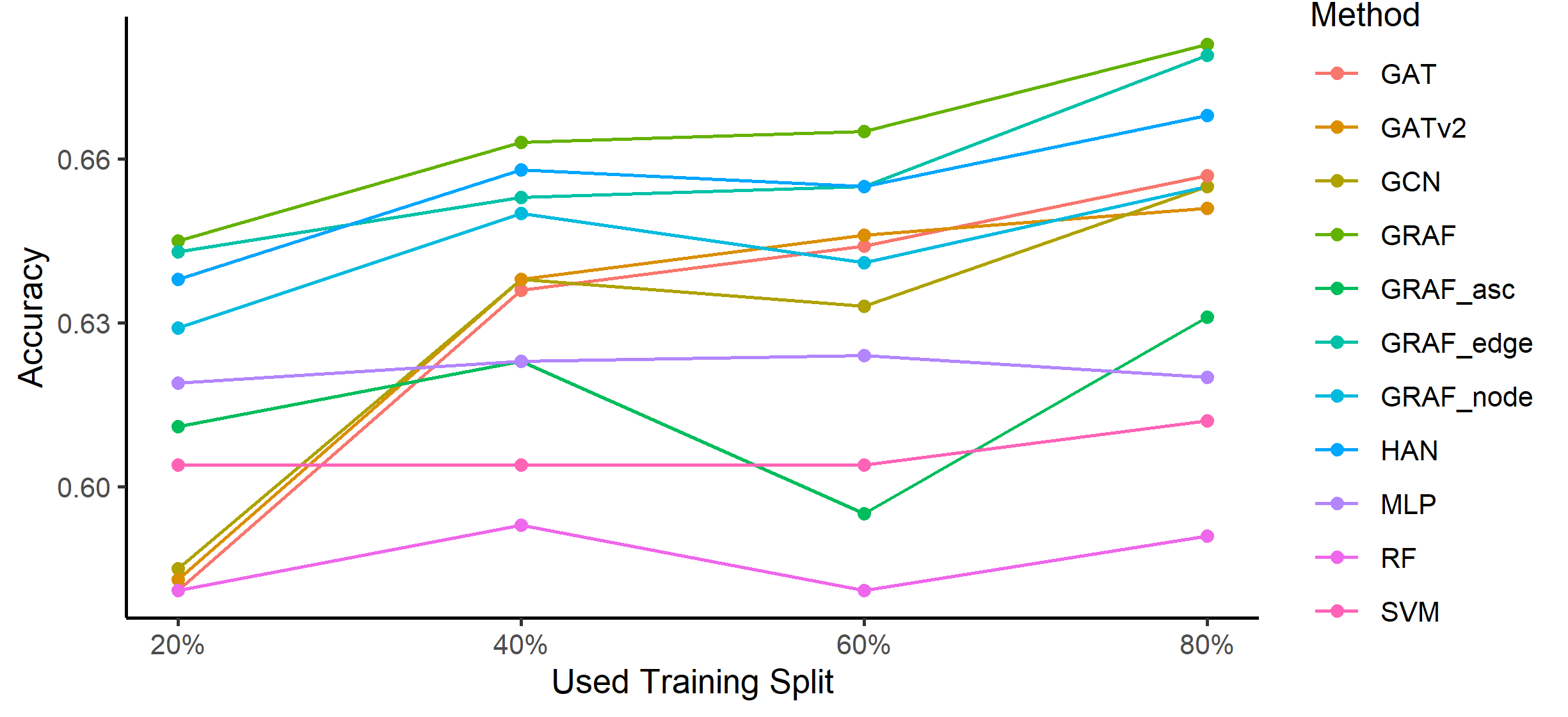}
  \caption{Performance with different training splits on IMDB data (accuracy).}
 \label{gfig:s1_3}
\end{figure}

\begin{figure}[h]
  \centering
  \includegraphics[width=0.9\linewidth]{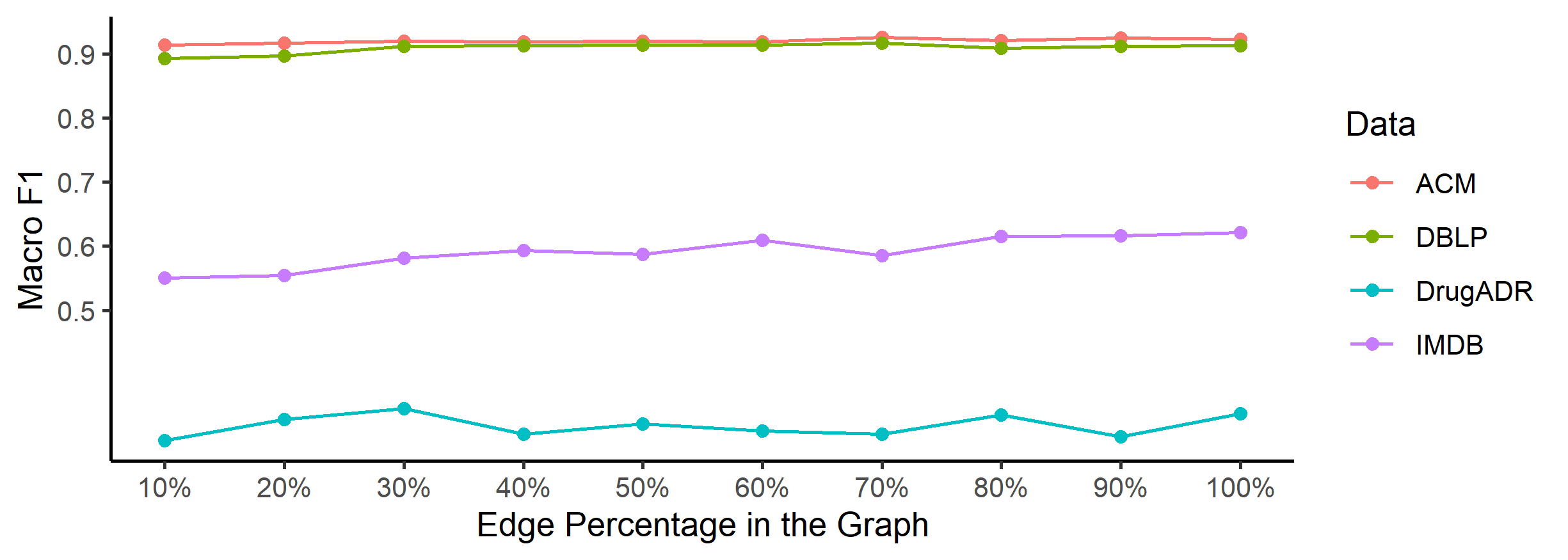}
 \caption{Performance with different edge percentages (macro F1).}
 \label{gfig:s2}
\end{figure}

\begin{figure}[h]
  \centering
  \includegraphics[width=0.9\linewidth]{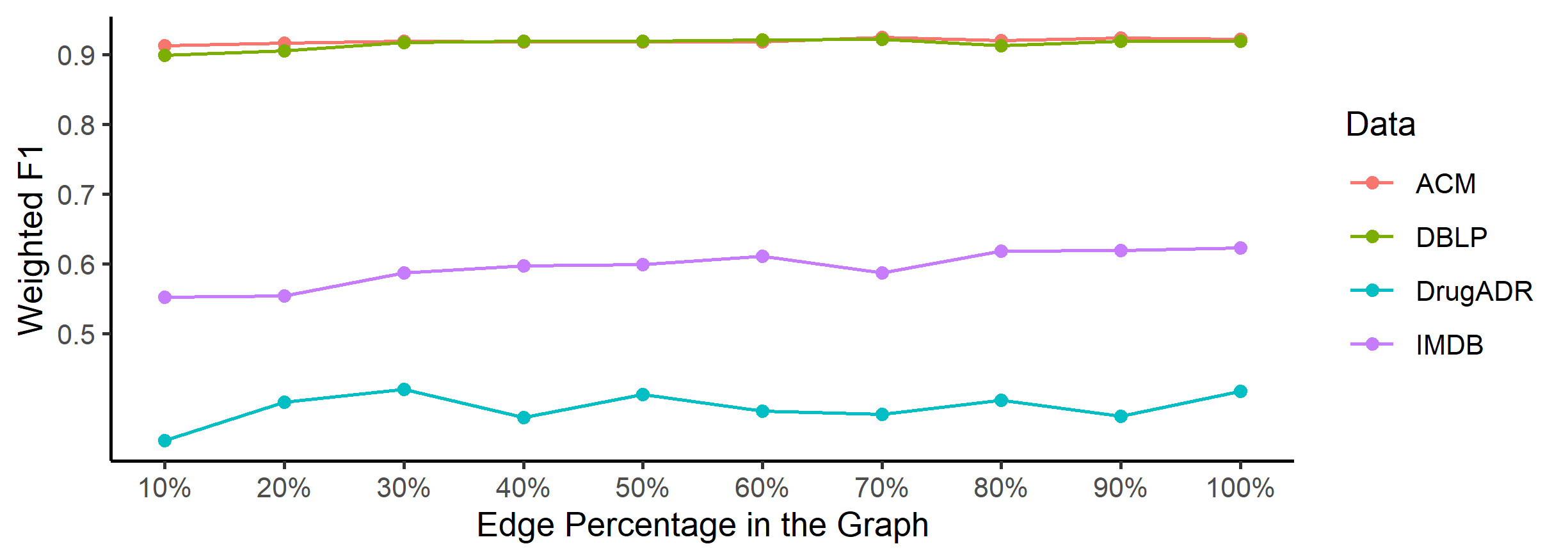}
  \caption{Performance with different edge percentages (weighted F1).}
 \label{gfig:s3}
\end{figure}

\begin{figure}[!h]
  \centering
  \includegraphics[width=0.9\linewidth]{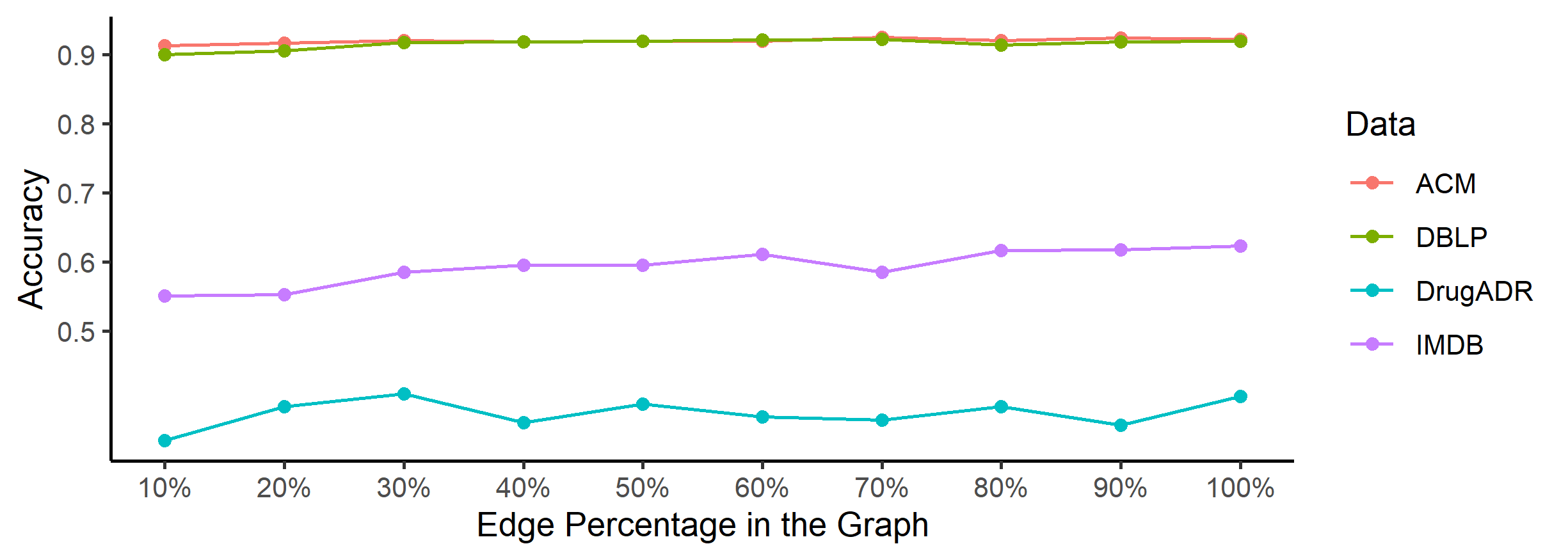}
  \caption{Performance with different edge percentages (accuracy).}
 \label{gfig:s4}
\end{figure}
\newpage

\bibliography{bib}
\bibliographystyle{unsrt}